\documentclass[letterpaper, 10 pt, conference]{ieeeconf}

\IEEEoverridecommandlockouts  

\usepackage{graphics} 
\usepackage{epsfig} 
\usepackage{mathptmx} 
\usepackage{times} 
\usepackage{amsmath} 
\usepackage{amssymb}  
\usepackage{color}
\usepackage[linesnumbered,ruled,vlined]{algorithm2e}

\newtheorem{definition}{Definition}
\newtheorem{thm}{Theorem}
\newtheorem{rem}{Remark}

\newtheorem{prop}{Proposition}
\newtheorem{problem}{Problem}
\newtheorem{spec}{Specification}
\SetAlgoVlined

\title{\LARGE \bf Probabilistically Safe Control of Noisy Dubins Vehicles}

\author{Igor Cizelj and Calin Belta
\thanks{This work was partially supported by 
the ONR  MURI under grant N00014-10-10952
and by the NSF under grant CNS-0834260. }
\thanks{The authors are with the Division of Systems Engineering at Boston University, Boston, MA 02215, USA. Email:
        {\tt\small $\{$icizelj,cbelta$\}$@bu.edu}.}%
}

\begin{document}
\maketitle
\thispagestyle{empty}
\pagestyle{empty}

\begin{abstract}
We address the problem of controlling a stochastic version of a Dubins 
vehicle such that the probability of satisfying a temporal logic specification over a set of properties at the regions in a partitioned environment is maximized. We assume that the vehicle can determine its precise initial position in a known map of the environment. However, inspired by practical limitations, we assume that the vehicle is equipped with noisy actuators and, during its motion in the environment, it can only measure its angular velocity using a limited accuracy gyroscope. Through quantization and discretization, we construct a finite approximation for the motion of the vehicle in the form of a Markov Decision Process (MDP). We allow for task specifications given as temporal logic statements over the environmental properties, and use tools in Probabilistic Computation Tree Logic (PCTL) to generate an MDP control policy that maximizes the probability of satisfaction. We translate this policy to a vehicle feedback control strategy and show that the probability that the vehicle satisfies the specification in the original environment is bounded from below by the maximum probability of satisfying the specification on the MDP.
\end{abstract}

\section{Introduction}
\label{sec:intro}
In ``classical'' motion planing problems \cite{Lavalle:planning}, the specifications are usually restricted to simple primitives of the type ``go from $A$ to $B$ and avoid obstacles", where $A$ and $B$ are two regions of interest in some environment. Often this is not rich enough to describe a task of interest in practical applications.
Recently, it has been shown that temporal logics, such as Linear Temporal Logic (LTL) and Computational Tree Logic (CTL), can serve as rich languages capable of specifying complex motion missions such as ``go to region $A$ and avoid region $B$ unless regions $C$ or $D$ are visited'' (see, for example, \cite{karaman:vehicle}, \cite{kloetzer:fully}, \cite{kress-gazit:whereswaldo?}, \cite{kavraki:motion2010}). 

In order to use tools from formal verification and automata games \cite{baier:principles} for motion planning and control, most of the works using temporal logics as specification languages assume that the motion of the vehicle in the environment can be modeled as a finite system. This usually takes the form of a finite transition system \cite{Clarke1999} that is either deterministic (applying an available action triggers a unique transition \cite{kloetzer:fully}) or nondeterministic (applying an available action can enable multiple transitions, with no information on their likelihoods \cite{KlBe-HSCC08-book}). More recent results show that,  
if sensor and actuator noise models can be obtained  through experimental trials, 
then the robot motion can be modeled as a Markov Decision Process (MDP), and 
probabilistic temporal logics, such as Probabilistic CTL (PCTL) and Probabilistic LTL (PLTL), can be used for motion planning and control (see \cite{LWAB10}).

However, robot dynamics are normally described by control systems with state and control variables evaluated over infinite domains. A widely used approach for temporal logic verification and control of such a system is through the construction of a finite abstraction \cite{tabuada:lineartime,Girard:approximately,YoBe-TAC-2011}). Even though recent works discuss the construction of abstractions for stochastic systems \cite{julius:approximations, Abate:Markov,D'Innocenzo:Approximate}, the existing methods are either not applicable to robot dynamics or are computationally infeasible given the size of the problem in most robotic applications. 

In this paper, we provide a conservative solution to the problem of controlling a stochastic Dubins vehicle such that the probability of satisfying a temporal logic specification over a set of properties at the regions in a partitioned environment is maximized. Inspired by a realistic scenario of an indoor vehicle leaving its charging station, we assume that the vehicle can determine its precise initial position in a known map of the environment. The actuator noise is modeled as a random variable with an arbitrary continuous probability distribution supported on a bounded interval. Also, we assume that the vehicle is equipped with a limited accuracy gyroscope, which measures its angular velocity, as the only means of measurement available. 

By discretization and quantization, we capture the motion of the vehicle as well as the position uncertainty as a finite state MDP. In this setup, the vehicle control problem is converted to the problem of finding a control policy for an MDP such that the probability of satisfying a PCTL formula is maximized. For the latter, we use the approach from \cite{LWAB10}. By establishing a mapping between the states of the MDP and the sequences of measurements obtained from the gyroscope, we show that a policy for the MDP becomes equivalent to a feedback control strategy for the vehicle in the environment. Finally, we show that the probability that the vehicle satisfies the specification in the original environment is bounded from below by the maximum probability of satisfying the specification on the MDP.

The main contribution of this work  lies in the application, since the result holds for a realistic vehicle with noisy actuators and a limited accuracy gyroscope. The method that we propose here is closely related to ``classical" Dynamic Programming (DP) - based approaches \cite{alterovitz:stochastic}. In these problems, the set of allowed specifications is restricted to reaching a given destination state, whereas our PCTL control framework allows for richer, temporal logic specifications and multiple destinations. 
In \cite{shkolnik:reachability}, the authors solve the problem of reaching a given destination while avoiding obstacles using the Rapidly-exploring Random Tree (RRT) algorithm that takes into account local reachability, as defined by differential constraints. In our approach, in order to obtain an MDP, a tree is also constructed (see Sec. \ref{section:reachability}), but we take into account reachability under uncertainty. Moreover, our approach produces a feedback control strategy and a lower bound on the probability of satisfaction, whereas the method presented in \cite{shkolnik:reachability}  only returns a collision-free trajectory. In addition, these methods differ from our work since they require precise state of the vehicle, at all times, whereas in our case, it is always uncertain. 

The remainder of the paper is organized as follows. In Sec. \ref{sec:prelim}, we introduce the necessary notation and review some preliminary results. 
We formulate the problem and outline the approach in Sec. \ref{sec:problem}. 
The discretization and quantization processes leading to the construction of the MDP model are described in Secs.  \ref{sec:approximation}, \ref{sec:uncertainty}, and \ref{sec:MDP}. The vehicle control policy is obtained in Sec. \ref{sec:control}. Case studies illustrating our approach are presented in Sec. \ref{sec:casestudy}. We conclude with final remarks and directions for future work in Sec. \ref{sec:conclusion}.

\section{Preliminaries}
\label{sec:prelim}

In this section, we provide a short and informal introduction to Markov Decision Processes (MDP) and Probabilistic Computation Tree Logic (PCTL). For details, the reader is referred to \cite{baier:principles}. 

\begin{definition}[MDP]
\label{def:MDP}
A labeled MDP $M$ is a tuple $(S,s_0,Act,A, P,\Pi,h)$, where 
$S$ is a finite set of states; $s_0 \in S$ is the initial state;  $Act$ is a finite set of actions; $A: S\rightarrow 2^{Act}$ is a function specifying the enabled actions at a state $s$; $P: S \times Act \times S \rightarrow [0,1]$ is a transition probability function such that for all states $s \in S$ and actions $a \in A(s)$: $\sum_{s' \in S}P (s,a,s')=1$, and for all actions $a\notin A(s)$ and $s'\in S$, $P (s,a,s')=0$; $\Pi$ is the set of propositions; and $h: S \rightarrow 2^{\Pi}$ is a function that assigns some propositions in $\Pi$ to each state of $s\in S$.
\end{definition}

A path $\omega$ through an MDP is a sequence of states $\omega=s_0s_1\ldots s_is_{i+1}\ldots$, where each transition is induced by a choice of an action at the current step $i$. We denote the set of all finite paths by $\text{Path}^{fin}$ (the MDP will be clear from the context).

\begin{definition}[MDP Control Policy]
\label{def:ConPol}
A control policy $\mu$ of an MDP $M$ is a function $\mu : \text{Path}^{fin} \rightarrow Act$ that specifies the next action to be applied 
after every finite path. 
\end{definition}

Probabilistic Computational Tree Logic (PCTL) is a probabilistic extension of CTL that includes the probabilistic operator $\mathcal{P}$.  
Formulas of PCTL are constructed by connecting propositions from a set $\Pi$ using Boolean operators ($\neg$ (negation), $\wedge$ (conjunction), and $\rightarrow$ (implication)), temporal operators ($\bigcirc$ (next), $\mathcal{U}$ (until)), and the probabilistic operator $\mathcal{P}$. 
For example, formula $\mathcal{P}_{max=?}[\neg \pi_3 \:  \mathcal{U}   \pi_4 ]$ asks for the maximum probability of reaching the states of an MDP satisfying $\pi_4$, without passing through states satisfying $\pi_3$.
The more complex formula $\mathcal{P}_{max=?} [\neg \pi_3 \, \mathcal{U} \, (\pi_4 \wedge \mathcal{P}_{\geq 0.5}[\neg \pi_3 \, \mathcal{U} \, \pi_1])$] asks for the maximum probability of eventually visiting states satisfying $\pi_4$ and then with probability greater than $0.5$ states satisfying $\pi_1$, while always avoiding states satisfying $\pi_3$.
Probabilistic model-checking tools, such as PRISM (see \cite{kwiatkowska:probabilistic}), can be used to find these probabilities. 
Simple adaptations of the model checking algorithms, such as the one presented in \cite{LWAB10}, can be used to find 
the corresponding control policies.


\section{Problem Formulation and Approach}
\label{sec:problem}
A Dubins vehicle (\cite{Dubins:Oncurves}) is a unicycle with constant forward speed and bounded turning radius moving in a plane. In this paper, we consider a stochastic version of a Dubins vehicle, which captures actuator noise:
\begin{equation}
\begin{bmatrix}
 \dot x\\  \dot y\\ \dot \theta
\end{bmatrix}
= 
\begin{bmatrix}
\cos(\theta)\\ 
\sin(\theta)\\ 
u+\epsilon
\end{bmatrix}, \text{ } u  \in U,
\label{eq:system2}
\end{equation}
where $(x,y) \in \mathbb{R}^2$ and $\theta \in [0,2 \pi)$ are the position and orientation of the vehicle in a world frame, 
$u$ is the control input, $U$ is the control constraint set, and $\epsilon$ is a random variable modeling the actuator noise. We assume that $\epsilon$ has an arbitrary continuous probability distribution supported on the bounded interval $[-\epsilon_{max},\epsilon_{max}]$.
The forward speed is normalized to $1$ and $\rho$ is the minimum turn radius.
We denote the state of the system by $q=[x,y,\theta]^T \in SE(2)$.

As it will become clear later, the control strategy proposed in this paper works for any finite set of controls $U$. However, 
motivated by the fact that the optimal Dubins paths use only three inputs (\cite{Dubins:Oncurves}), we assume 
\[
U = \{-1/\rho,0,1/\rho\}.
\]
We define 
\[
W=\{u+\epsilon | u \in U, \epsilon \in [-\epsilon_{max},\epsilon_{max}] \}
\]
as the set of applied control inputs, i.e, the set of angular velocities that are applied to the system in the presence of noise. We assume that time is uniformly discretized (partitioned) into stages (intervals) of length $\Delta t$, where stage $k$ is from $(k-1)\Delta t$ to $k \Delta t$. The duration of the motion is finite and it is denoted by  $K \Delta t$.~\footnote[1]{Since PCTL has infinite time semantics, after $K \Delta t$ the system remains in the state achieved at $K \Delta t$. In Remark 1 (Sec. \ref{sec:motion}), we explain how  to determine $K$.} We denote the control input and the applied control input at stage $k$ as $u_k \in U$ and $w_k \in W$, respectively. 

We assume that the noise $\epsilon$ is piece-wise constant, i.e, it can only change at the beginning of a stage. This assumption is motivated by practical applications, in which a servo motor is used as an actuator for the turning angle (see e.g., \cite{Mazo04robustarea}). This implies that the applied control is also piece-wise constant, i.e., $w:[(k-1)\Delta t,k\Delta t] \rightarrow W$, $k=1,\ldots,K$, is constant over each stage. We assume that the vehicle is equipped with only one sensor, which is a limited accuracy gyroscope. At stage $k$, this returns the measured interval $[\underline{w}_k,\overline{w}_k] \subset [u_k-\epsilon_{max},u_k+\epsilon_{max}]$ containing the applied control input.

The vehicle moves in a planar environment that is partitioned$\footnote[2]{Throughout the paper, we relax the notion of a partition by allowing regions to share bundaries.}$ into a set of polytopic regions $R$. Let $R_{unsafe} \subset R$ denote a set of unsafe regions, and $R_{safe} = R \setminus R_{unsafe}$ define the set of safe regions. One set of regions $R_{p} \subset R_{safe}$ is labeled with ``pick-up", and another set $R_d \subset R_{safe}$ is labeled with ``drop-off". In this work, we assume that the motion specification is as follows: 

\begin{spec}\label{spec:main}
``Starting from an initial state $q_{init}$, the vehicle is required to reach a pick-up region to pick up a load. Then, the vehicle should go to a drop-off region to drop off the load. At all times, the vehicle should avoid the unsafe regions.''
\end{spec}

Let $\Pi=\{\pi_p,\pi_d,\pi_u\}$ be a set of propositions, where $\pi_p,\pi_d$, and $\pi_u$ label the pick-up, drop-off, and unsafe regions, respectively. We define $[\pi_p]=\{(x,y) \in \mathbb{R}^2|(x,y) \in \cup_{r \in R_p} r\}$, $[\pi_d]=\{(x,y) \in \mathbb{R}^2|(x,y) \in \cup_{r \in R_d} r\}$ and $[\pi_u]=\{(x,y) \in \mathbb{R}^2|(x,y) \in \cup_{r \in R_{unsafe}} r\}$ as the set of all positions that satisfy propositions $\pi_p$, $\pi_d$, and $\pi_u$, respectively.


We assume that the vehicle can precisely determine its initial state $q_{init}=[x_{init},y_{init},\theta_{init}]^T$ in a known map of the environment. While the vehicle moves, gyroscope measurements $[\underline{w}_k,\overline{w}_k] $ are available at each stage $k$. We define a {\it vehicle control strategy} as a map that takes as input a sequence of measured intervals $[\underline{w}_1,\overline{w}_1][\underline{w}_2,\overline{w}_2]\ldots[\underline{w}_{k-1},\overline{w}_{k-1}]$ 
and returns the control input $u_k \in U$ at stage $k$. We are ready to formulate the main problem that we consider in this paper:

\begin{problem}\label{problem:main}
Given a partitioned environment $R$, a vehicle model described by Eqn. (\ref{eq:system2}) with initial state $q_{init}$, a motion task in the form of Specification \ref{spec:main}, find a vehicle control strategy that maximizes the probability of satisfying the specification.
\end{problem}

The requirement that the vehicle maximizes the probability of satisfying Specification \ref{spec:main} translates to the following PCTL formula:
\begin{equation}
\phi: \mathcal{P}_{max=?}[\lnot \pi_u \mathcal{U} (\lnot \pi_u \wedge \pi_p \wedge
 \mathcal{P}_{>0}[\lnot \pi_u \mathcal{U}(\lnot \pi_u \wedge \pi_d)])].
\label{eq:phi}
\end{equation}
To fully specify Problem \ref{problem:main}, we need to define the satisfaction of a PCTL formula $\phi$ by a trajectory $q:[0,K\Delta t] \rightarrow SE(2)$ of the system from Eqn. (\ref{eq:system2}). For $\Theta \in 2^{\Pi}$, let $[\Theta]$ be the set of all positions in $\mathbb{R}^2$ satisfying all and only propositions $\pi \in \Theta$.
The word corresponding to a state trajectory $q(t)$ is a sequence $o=o_1o_2o_3\dots$, $o_k \in 2^{\Pi}$, $k \geq1$, generated 
according to the following rules, for all $t \in [0,K \Delta t]$ and $k \in \mathbb{N}$, $k \geq 1$: (i) $(x(0),y(0)) \in [o_1]$; (ii) if $(x(t),y(t)) \in [o_k]$ and $o_k \neq o_{k+1}$, then $\exists$ $t' \geq t$ s.t. a) $(x(t'),y(t')) \in [o_{k+1}]$ and b) $(x(\tau),y(\tau)) \notin [\pi]$, $\forall \tau \in [t,t']$, $\forall \pi \in \Pi \setminus (o_k \cup o_{k+1})$; (iii) if $(x(K \Delta t),y(K \Delta t)) \in [o_k]$ then $o_i=o_k$ $\forall i \geq k$. 
Informally, the word produced by $q(t)$ is the sequence of sets of satisfied propositions as time evolves. A trajectory $q(t)$ satisfies PCTL formula $\phi$ if and only if the word generated according to the rules stated above satisfies the formula.

In this paper, we develop an approximate solution to Problem \ref{problem:main} consisting of three steps. First, by discretizing the noise interval, we define a finite subset of the set of possible applied control inputs. We use this to define a Quantized System (QS) that approximates the original system given by Eqn.  (\ref{eq:system2}). Second, we capture the uncertainty in the position of the vehicle and map the QS to 
an MDP. Finally, we find a control policy for the MDP that maximizes the probability of satisfying the specification, and translate this policy to a vehicle control strategy.
In addition, we show that the probability that the original system under the obtained control strategy satisfies $\phi$ is bounded from bellow by the obtained maximum probability. 


\section{Approximation}
\label{sec:approximation}
\subsection{Quantized System}
\label{sec:QS}

We use $q_k(t)$ and $w_k$, $t \in [(k-1)\Delta t, k\Delta t]$, $k=1,\ldots,K$ to denote the state trajectory and the constant applied control at stage $k$, respectively. 
With a slight abuse of notation, we use $q_k$ to denote the end of state trajectory $q_k(t)$, i.e., $q_k=q_k(k \Delta t)$. Given a state $q_{k-1}$, the state trajectory $q_k(t)$ can be derived by integrating the system given by Eqn. (\ref{eq:system2}) from the initial state $q_{k-1}$, and taking into account that the applied control is constant and equal to $w_k$. 
Throughout the paper, we will also denote this trajectory by $q_k(q_{k-1},w_k,t)$, when we want to explicitly capture the initial state $q_{k-1}$ and the constant applied control $w_k$.

Motivated by practical applications, we assume that the measurement resolution of the gyroscope, i.e., the length of $[\underline{w}_k,\overline{w}_k]$, is constant, and we denote it by 
$\Delta \epsilon$. For simplicity of presentation, we also assume that $n\Delta \epsilon=2\epsilon_{max}$, for some $n \in \mathbb{Z}^+$. However, the following approach works as long as $\Delta \epsilon \leq 2 \epsilon_{max}$. Let $\underline{\epsilon}_i=-\epsilon_{max}+(i-1) \Delta \epsilon$ and $\overline{\epsilon}_i=-\epsilon_{max}+i \Delta \epsilon$, $i=1,\ldots,n$. Then, $[-\epsilon_{max},\epsilon_{max}]$ can be partitioned$\footnote[3]{Throughout the paper, we relax the notion of a partition by allowing the endpoints of the intervals to overlap.}$ into $n$ intervals: $[\underline{\epsilon}_i,\overline{\epsilon}_i]$, $i=1,\ldots,{n}$. 

For each interval we define a representative value $\epsilon_i=\frac{\underline{\epsilon}_i+\overline{\epsilon}_i}{2}$, $i=1,\ldots,n$, i.e., $\epsilon_i$ is the midpoint of interval $[\underline{\epsilon}_i,\overline{\epsilon}_i]$. We denote the set of all noise intervals and the set of their representative values as $\mathcal{E}=\{[\underline{\epsilon}_1,\overline{\epsilon}_1],\ldots,[\underline{\epsilon}_{n},\overline{\epsilon}_{n}]\}$ and $E=\{\epsilon_1,\ldots,\epsilon_{n}\}$, respectively. 
At stage $k$, the gyroscope returns the measured interval $[u_k-\underline{\epsilon}_k,u_k+\overline{\epsilon}_k]$ containing the applied control input, where $[\underline{\epsilon}_k,\overline{\epsilon}_k] \in \mathcal{E}$. Note that, since $u_k \in U$ is known, we can obtain obtain the noise interval $[\underline{\epsilon}_k,\overline{\epsilon}_k] \in \mathcal{E}$, containing the noise at stage $k$, from the measured interval.


Recall that $\epsilon$ is a random variable with an arbitrary continuous probability distribution supported on the bounded interval $[-\epsilon_{max},\epsilon_{max}]$. In this paper, we assume uniform distribution, but our approach is general in the sense that it holds for any continuous distribution supported on a bounded interval. For the uniform distribution the following holds: 
\begin{equation}
\text{Pr}(\epsilon \in [\underline{\epsilon}_i,\overline{\epsilon}_i])=\frac{|\overline{\epsilon}_i-\underline{\epsilon}_i|}{2 \epsilon_{max}}=\frac{\Delta \epsilon}{2\epsilon_{max}}=\frac{1}{n},
\label{eq:probability}
\end{equation}
$[\underline{\epsilon}_i,\overline{\epsilon}_i] \in \mathcal{E}$,  $i=1,\ldots,n$. 

We define $W_d=\{u+\epsilon \text{ } | \text{ } u \in U,  \epsilon \in E\} \subset W$ as a finite set of applied control inputs. Also, let $\omega: U \rightarrow W_d$ be a random variable, where $\omega(u)=u + \epsilon $ with the probability mass function $p_{\omega}(\omega(u)=u+\epsilon)=\frac{1}{n}$, $\epsilon \in E$. This probability follows from Eqn. (\ref{eq:probability}) since $\epsilon \in E$ is the representative value of interval $[\underline{\epsilon},\overline{\epsilon}] \in \mathcal{E}$. Finally, we define a Quantized System (QS) that approximates the original system as follows: The set of applied control inputs in QS is $W_d$; for a state $q_{k-1}$ and a control input $u_{k} \in U$, QS returns  
\begin{equation}
\label{eq:system5}
q_k(q_{k-1},\omega(u_k),t)=q_k(q_{k-1},u_k+\epsilon,t)
\end{equation}
with probability $\frac{1}{n}$, where $\epsilon \in E$.
Since for $u_{k} \in U$ the applied control input is $u_{k}+\epsilon \in W_d$ with probability $\frac{1}{n}$, the returned state trajectory at stage $k$ is $q_k(q_{k-1},u_{k}+\epsilon,t)$ with probability $\frac{1}{n}$.

\subsection{Reachability graph}
\label{section:reachability}

\begin{figure}[htb]
\begin{center}
\includegraphics[width=0.44\textwidth]{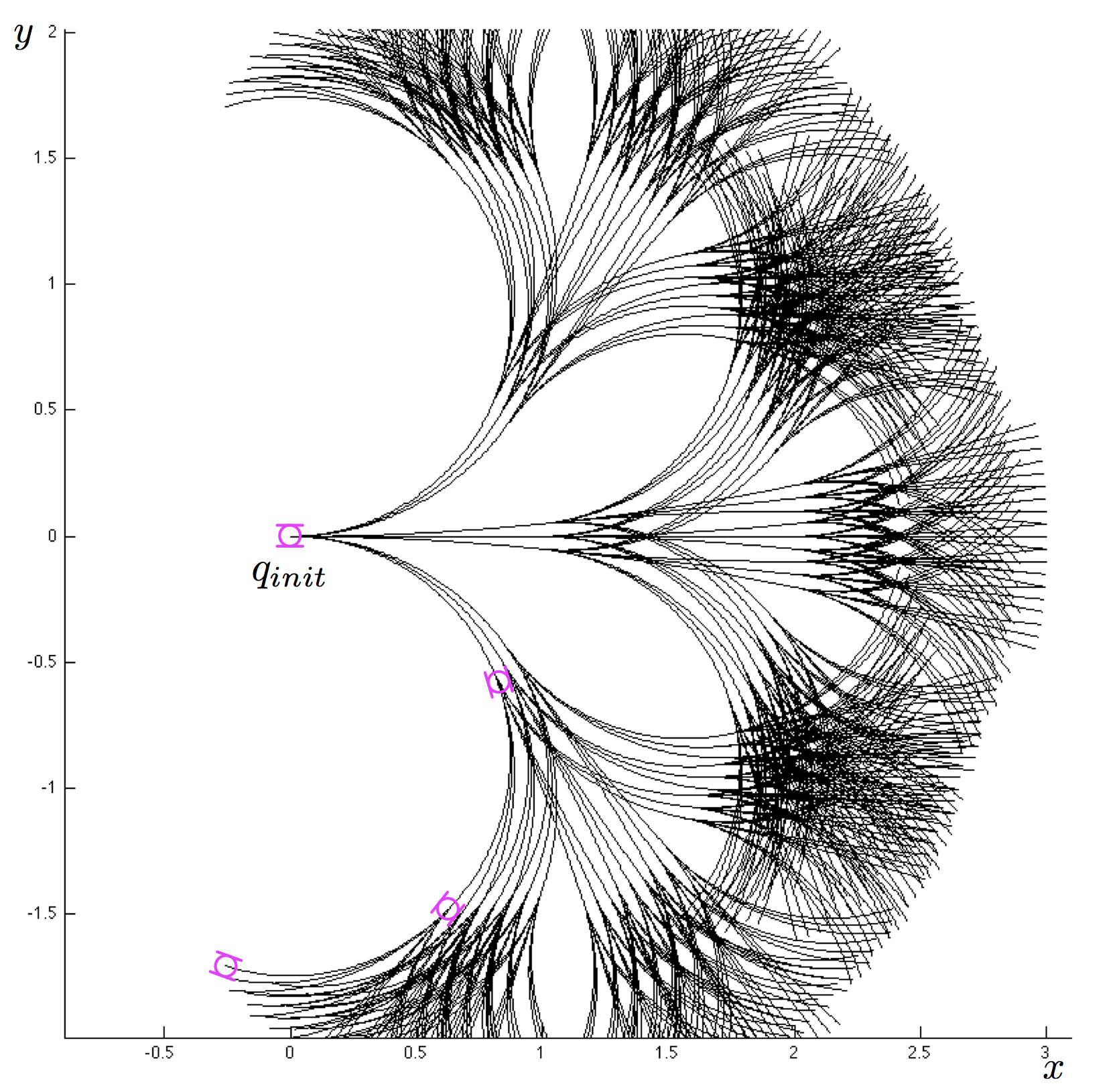}
\end{center}
\caption{The projection of reachability graph $G_3(q_{init})$ in $\mathbb{R}^2$ when $U=\{-\frac{\pi}{3},0,\frac{\pi}{3}\} $ and $E=\{-0.1,0,0.1\}$ with $\Delta t=1.2$. Magenta objects represent the states of the vehicle. }
\label{fig:reachabilitygraph}
\end{figure}

We denote $u_1u_2 \ldots u_K$, in which each $u_k \in U$ gives a control input at stage $k$, as a finite sequence of control inputs of length $K$. 
We use $\Sigma_K$ to denote the set of all such sequences.
For the initial state $q_{init}$ and $\Sigma_K$, we define the reachability graph $G_K(q_{init})$ (see \cite{Lavalle:planning} for a related definition), which encodes the set of all state trajectories originating from $q_{init}$ that can be obtained, with a positive probability, by applying sequences of control inputs from $\Sigma_K$ according to QS given by Eqn. (\ref{eq:system5}). In Fig. \ref{fig:reachabilitygraph} we give an example of a reachability graph.

Note that, by using a gyroscope with a finer measurement resolution (i.e., by decreasing $\Delta \epsilon$), a more dense reachability graph can be obtained. In the theoretical limit, as $\Delta \epsilon \rightarrow 0$, $G_K(q_{init})$ approaches the set of all trajectories, originating from $q_{init}$, that can be generated by the original system.

\section{Position Uncertainty}
\label{sec:uncertainty}
Since the specification (see Eqn. (\ref{eq:phi})) is a statement about the propositions satisfied by the regions in the partitioned environment, in order to answer whether some state trajectory satisfies PCTL formula $\phi$, it is sufficient to know its projection in $\mathbb{R}^2$. Therefore, we focus only on the position uncertainty.

The position uncertainty of the vehicle when its nominal position is $(x,y) \in \mathbb{R}^2$ is modeled as a disc centered at $(x,y)$ with radius $\xi \in \mathbb{R}$, where $\xi$ denotes the uncertainty:
\begin{equation}
D((x,y),\xi)=\{(x',y') \in \mathbb{R}^2|||(x,y),(x',y')|| \leq \xi\},
\end{equation} 
where $|| \cdot ||$ denotes the Euclidian distance. Next, we explain how to obtain $\xi$. 

\begin{figure}[htb]
\begin{center}
\includegraphics[width=0.475\textwidth]{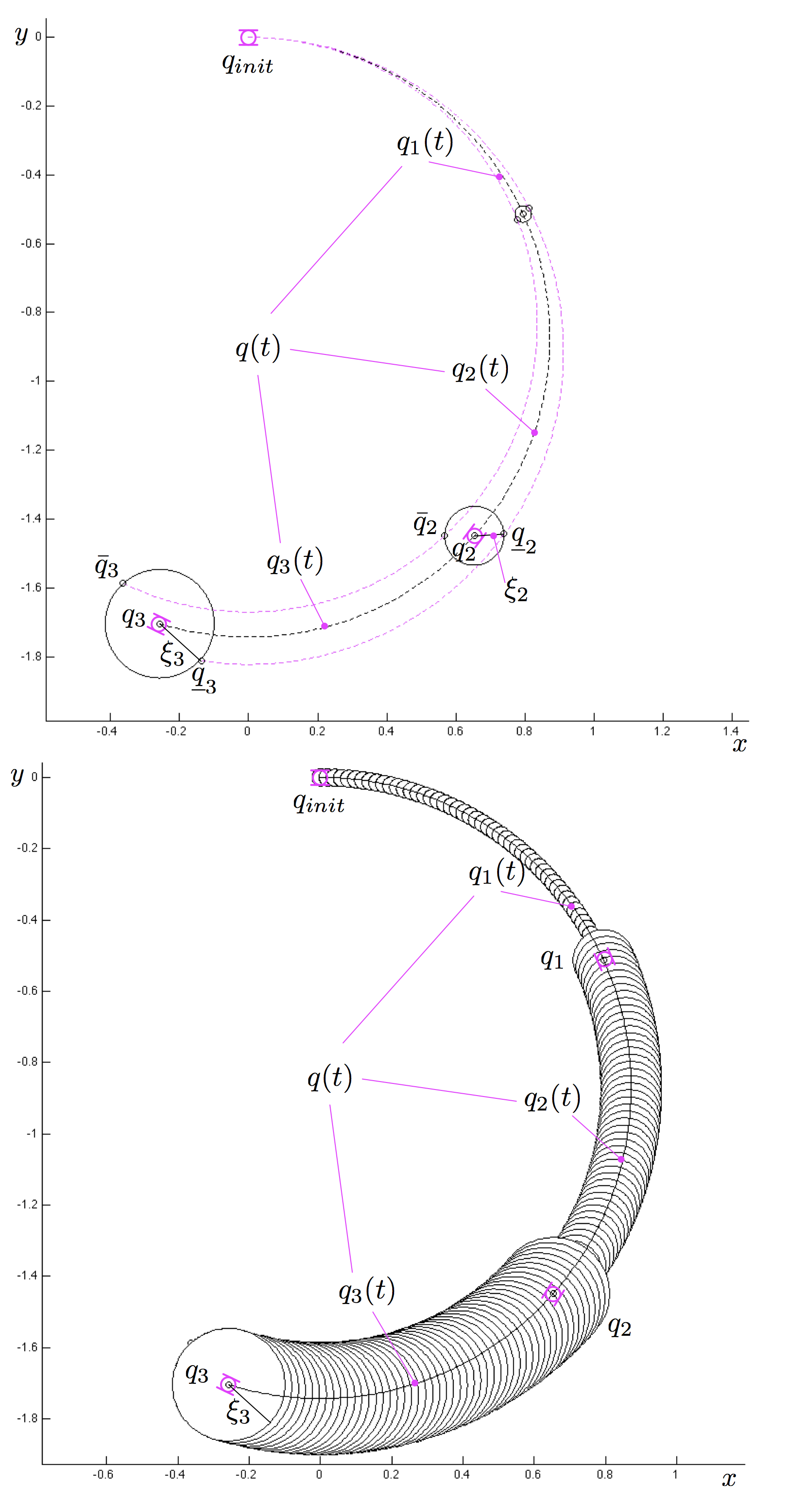}
\end{center}
\caption{Above: Evolution of the position uncertainty along the state trajectory $q(t)$, where $q(t)$ is partitioned into $3$ state trajectories, $q_k(t)$, $k=1,2,3$. Below: The conservative approximation of region $D((x(t),y(t)),\xi(t))$ along the state trajectory $q(t)=[x(t),y(t),\xi(t)]^T$, when the uncertainty trajectory is $\xi(t')=\xi_k(t)$, $t' \in [(k-1)\Delta t,k\Delta t]$, where $\xi_k(t)=\xi_k$, $k=1,2,3$.  }
\label{fig:uncertaintyevolution}
\end{figure}

Any state trajectory $q(t) \in G_K(q_{init})$, $t \in [0, K\Delta t]$, can be partitioned into $K$ state trajectories: $q_k(t)=q(t')$, $t' \in [(k-1)\Delta t,k \Delta t]$, $k=1,\ldots,K$ (see  Fig. \ref{fig:uncertaintyevolution}). We denote the uncertainty at state $q_k$ as $\xi_k$. Let $u_k+\epsilon_k \in W_d$  be the applied control input at stage $k$ such that $q_{k}(t)=q_k(q_{k-1},u_{k}+\epsilon_{k},t)$, $k=1,\ldots,K$, with $q_0=q_{init}$. Then, we set the uncertainty at state $q_k=[x_k,y_k,\theta_k]^T$ equal to: 
\begin{equation}
\begin{split}
\xi_k=\operatorname*{max}_{[x',y',\theta']^T \in \{\overline{q}_{k},\underline{q}_{k}\}} \{||(x_k,y_k),(x',y'))||\} \text{  where }\\
\label{eq:unceratiny}
\underline{q}_{k}(t)=\underline{q}_k(\underline{q}_{k-1},u_k+\underline{\epsilon}_k,t) \text{ and }
\overline{q}_{k}(t)=\overline{q}_k(\overline{q}_{k-1},u_k+\overline{\epsilon}_k,t),
\end{split}
\end{equation} 
for $k=1, \ldots,K$, where $\underline{q}_0=\overline{q}_0=q_{init}$. 

Eqn. (\ref{eq:unceratiny}) is obtained using a worst case scenario assumption. 
If $u_k + \epsilon_k \in W_d$ is the applied control input for QS,  
the corresponding applied control input at stage $k$ for the original system is in 
$[u_k-\underline{\epsilon_k},u_k+\overline{\epsilon}_k]$, where $\epsilon_k \in [\underline{\epsilon_k},\overline{\epsilon}_k] \in \mathcal{E}$. 
The position of the end state of the original system at stage $k$ would be the 
farthest (in the Euclidean sense) from $q_k$, if the applied control input was either $u_i+\underline{\epsilon}_i$, $i=1,\ldots,k$, or $u_i+\overline{\epsilon}_i$, $i=1,\ldots,k$  (see \cite{Fraichard98pathplanning} for more details). An example is given in Fig. \ref{fig:uncertaintyevolution}. 

From Eqn. (\ref{eq:unceratiny})  it follows that, given a state trajectory $q(t) \in G_K(q_{init})$, $t \in [0, K\Delta t]$, the uncertainty is increasing as a function of time. The way the uncertainty changes along $q(t)$ makes it difficult to characterize the exact shape of the position uncertainty region. Instead, we use a conservative approximation of the region. We define $\xi : [0,K \Delta t] \rightarrow \mathbb{R}$ as an approximated uncertainty trajectory and we set $\xi(t)=\xi_k$, $t \in [(k-1)\Delta t,k \Delta t]$, $k=1,\ldots,K$, i.e., we set the uncertainty along the state trajectory $q_k(t)$ equal to the maximum value of the uncertainty along $q_k(t)$, which is at state $q_k$. An example illustrating this idea is given in Fig. \ref{fig:uncertaintyevolution}. 

In this work, we assume that the forward speed is constant and normalized to $1$. The uncertainty model presented above can be extended to take into account forward speed uncertainty as shown in \cite{Fraichard98pathplanning}.

\section{Construction of an MDP Model}
\label{sec:MDP}
\subsection{Satisfying $\phi$ under uncertainty}
\label{sec:motion}

Based on the rules presented in Sec. \ref{sec:problem}, to guarantee that a state trajectory $q(t) \in G_K(q_{init})$ satisfies $\phi$ (Eqn. (\ref{eq:phi})) when the uncertainty trajectory is $\xi(t)$, $t \in [0,K \Delta t]$, the following conditions need to be satisfied: (i) $D((x(t),y(t)),\xi(t)) \subseteq [\pi_p]$ for some $t \in [0, K\Delta t]$, (ii) $D((x(K \Delta),y(K \Delta)),\xi(K\Delta)) \subseteq [\pi_d]$ and (iii) $D((x(t),y(t)),\xi(t)) \cap [\pi_u]= \emptyset$ for all $t \in [0, K\Delta t]$. If satisfied, these conditions guarantee that a pickup region is entered, the end state is inside a drop-off region, and $R_{unsafe}$ is not entered along $q(t)$ when the uncertainty trajectory is $\xi(t)$.

Assume $q(t)$ and $\xi(t)$ are partitioned into $K$ state and uncertainty trajectories, respectively, s.t. $q_k(t)=q(t')$ and $\xi_k(t)=\xi(t')$, $t' \in [(k-1)\Delta t,k\Delta t]$, $k=1,\ldots,K$.
Then, the conditions stated above can be written as: (i) $D((x_k(t),y_k(t)),\xi_k(t)) \subseteq [\pi_p]$ for some $t \in [(k-1)\Delta t, k\Delta t]$ and some $k$, (ii) $D((x_K,y_K),\xi_K) \subseteq [\pi_d]$, and (iii) $D((x_k(t),y_k(t)),\xi_k(t)) \cap [\pi_u] = \emptyset$ for all $t \in [(k-1)\Delta t, k\Delta t ]$ and all $k$. Thus, by analyzing $q_k(t)$ when the uncertainty trajectory is $\xi_k(t)$, $k=1,\ldots,K$, we can answer if $q(t)$, when the uncertainty trajectory is $\xi(t)$, is satisfying. 

\begin{rem}
Note that the vehicle is subject to cumulative and unbounded position uncertainty. Even though there are results on how to overcome the cumulative nature of the position uncertainty (e.g., see  \cite{Fraichard98pathplanning}), in this work we assume that $q_{init}$ and the environment are such that $\exists K$ for which a satisfying state trajectory exists (if the assumption holds, then such a $K$ can always be found by using a gyroscope with a finer measurement resolution, which leads to a more dense reachability graph). If this assumption is violated (e.g., the minimum turn radius is too large to enter a pick-up region without entering $R_{unsafe}$), then there is no solution to the problem. For the rest of this paper, we use the smallest (the first) $K$ for which a satisfying state trajectory can be found. Using the smallest $K$ also reduces the computational complexity (see Sec. \ref{sec:control}).
\end{rem}

\subsection{MDP construction}

\begin{figure}[htb]
\begin{center}
\includegraphics[width=0.475\textwidth]{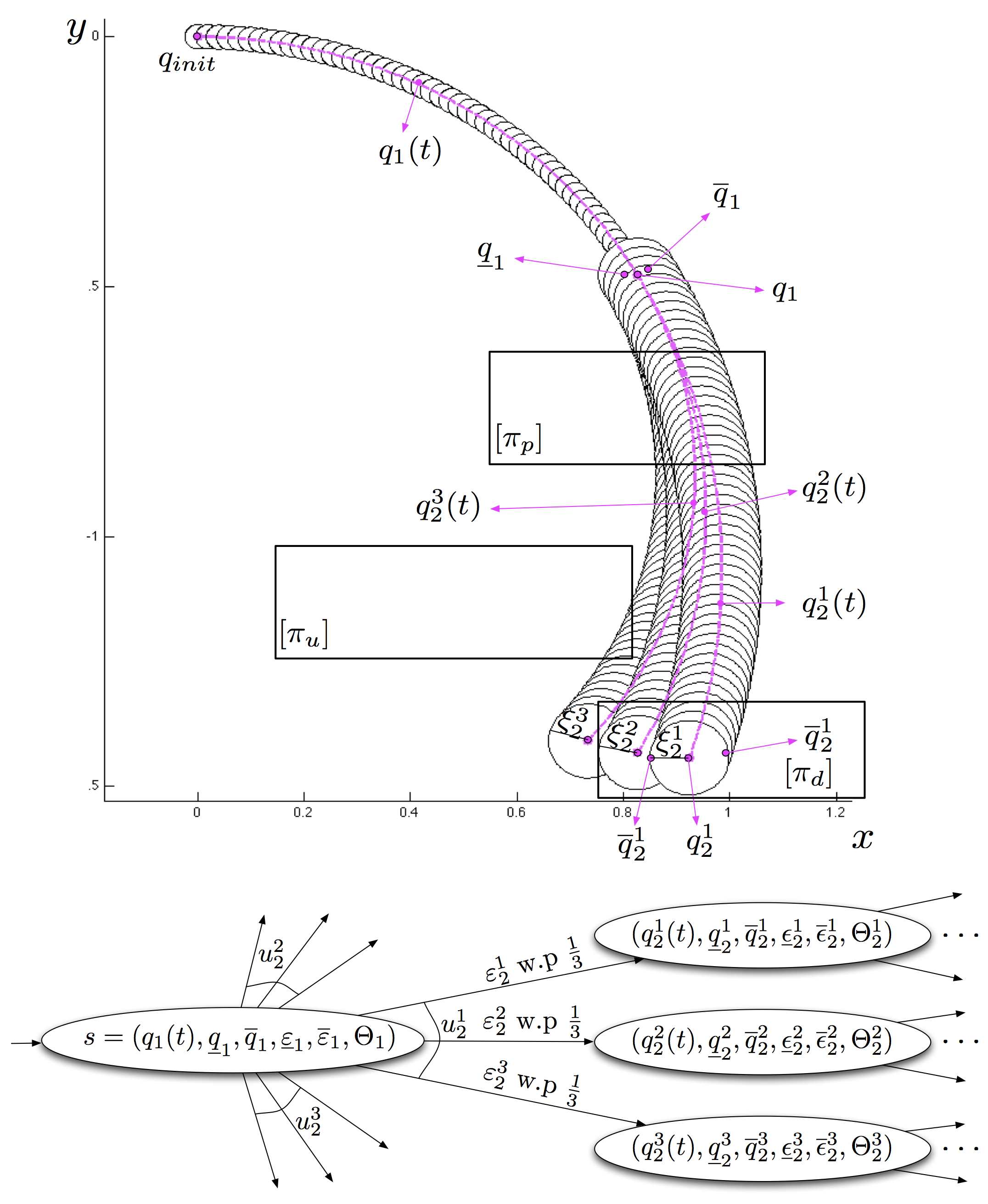}
\end{center}
\caption{Above: An example scenario corresponding to the MDP fragment shown below. For the state trajectory $q_2^1(t)=[x(t),y(t),\theta(t)]^T$, $t \in [\Delta t, 2\Delta t]$, when the uncertainty trajectory is $\xi_2^1(t)=\xi_2^1$ the following holds: (i) it can be guaranteed that a  pick-up region is entered (i.e., $\exists t \in [\Delta t, 2\Delta t]$ s.t. $D((x(t),y(t)),\xi(t)) \subseteq [\pi_p]$), (ii) the end state is inside of a drop-off region (i.e., $D((x(2 \Delta t),y(2 \Delta t)),\xi(2 \Delta t)) \subseteq [\pi_d]$), and (iii) $R_{unsafe}$ is not entered (i.e., $\forall t \in [\Delta t, 2\Delta t]$,  $D((x(t),y(t)),\xi(t)) \cap [\pi_u] = \emptyset$).
Thus, $\Theta_2^1=\{\pi_{p},\pi_{d}\}$. Similarly, $\Theta_2^2=\{\pi_{p},\pi_{d}\}$ but $\Theta_2^3=\{\pi_{p},\pi_u\}$. Below: A fragment of the MDP corresponding to the scenario shown above, where $[-\epsilon_{max}, \epsilon_{max}]$ is partitioned into $n=3$ intervals. Action $u_2^1 \in A(s)$ enables three transitions, each w.p. $\frac{1}{3}$. This corresponds to applied control input being equal to $u_2^1+\epsilon_2^i$ w.p. $\frac{1}{3}$, $\epsilon_2^i \in E$. The elements of the new states are: $q_2^i(t)=q_2^i(q_1,u_2^1+\epsilon_2^i,t)$; $\underline{q}_2^i(t)=\underline{q}_2^i(\underline{q}_1,u_2^1+\underline{\epsilon}_2^i,t)$; $\overline{q}_2^i(t)=\overline{q}_2^i(\overline{q}_1,u_2^1+\overline{\epsilon}_2^i,t)$; $[\underline{\epsilon}_2^i,\overline{\epsilon}_2^i] \in \mathcal{E}$ is s.t. $\epsilon_2^i \in  [\underline{\epsilon}_2^i,\overline{\epsilon}_2^i]$; and $\Theta_2^i$ is given above, where $i=1,2,3$. }
\label{fig:MDPGraphics}
\end{figure}

A labeled MDP $M$ that models the motion of the vehicle in the environment and the evolution of position uncertainty is defined 
as a tuple $(S,s_0,Act,A,P,\Pi,h)$ where:\\
$\bullet$ $S$ is the finite set of states. For every state trajectory  $q_k(t) \in G_K(q_{init})$, $t \in [(k-1)\Delta t,k\Delta t]$, $k=1,\ldots,K$, a state of the MDP is created. The meaning of the state is as follows: $(q(t),\underline{q},\overline{q},\underline{\epsilon},\overline{\epsilon},\Theta) \in S$ means that along the state trajectory $q(t)$, the uncertainty trajectory is
$$\xi(t)=\operatorname*{max}_{[x',y',\theta']^T \in \{\underline{q},\overline{q}\}}||(x,y),(x',y')||,$$
where $[x,y,\theta]^T$ is the end state of $q(t)$; The noise interval is $[\underline{\epsilon},\overline{\epsilon}] \in \mathcal{E}$;
For $\Theta \in 2^{\Pi}$: (i) $\pi_p \in \Theta$, (ii) $\pi_d \in \Theta$, and (iii) $\pi_u \in \Theta$, mean that (i) it can be guaranteed that a pick-up region is entered, (ii) it can be guaranteed that the end state is inside of a drop-off region, and (iii) it is possible to enter $R_{unsafe}$, along the state trajectory $q(t)$ when the uncertainty trajectory is $\xi(t)$ (see Fig. \ref{fig:MDPGraphics} for an example). Note that $\xi(t)$ is not an element of a state explicitly since it can be obtained from $q$, $\underline{q}$, and $\overline{q}$ (Eqn. (\ref{eq:unceratiny})).\\ 
$\bullet$ $s_0=(q_{init},q_{init},q_{init},\emptyset,\emptyset,\Theta_{init}) \in S$ is the initial state, where $\Theta_{init} \in 2^{\Pi}$ is the set of propositions satisfied at $q_{init}$.\\
$\bullet$ $Act=U \cup \varphi$ is the set of actions, where $\varphi$ is a dummy action; \\
$\bullet$ $A: S \rightarrow 2^{Act}$ gives the enabled actions at state $s$: at termination time, $A(s)=\varphi$, otherwise $A(s)=U$;\\
 $\bullet$ $P: S \times Act \times S \rightarrow [0,1]$ is a transition probability function (its construction is described below);\\
 $\bullet$ $\Pi = \{\pi_p,\pi_d,\pi_u\}$ is the set of propositions;\\
$\bullet$ $h: S \rightarrow 2^{\Pi}$ assigns propositions from $\Pi$ to states $s \in S$ according to the following rule: for $s=(q(t),\underline{q},\overline{q},\underline{\epsilon},\overline{\epsilon},\Theta)$, then $\pi_p \in h(s)$ iff $\pi_p \in \Theta$, $\pi_d \in h(s)$ iff $\pi_d \in \Theta$, and $\pi_u \in h(s)$ iff $\pi_u \in \Theta$.

\begin{algorithm}
{\small{
\label{alg:algorithm1}
\caption{Generating S and P}
\KwIn{$s \in S$, $u_k \in U$, $\epsilon_k \in E$, $S$, $P$}
\KwOut{$S$, $P$}
$(q_{k-1}(t),\underline{q}_{k-1},\overline{q}_{k-1},\underline{\epsilon}_{k-1},\overline{\epsilon}_{k-1},\Theta_{k-1})=s$;\\
$q_k(t)=[x_k(t),y_k(t),\theta_k(t)]^T=q_k(q_{k-1},u_k+\epsilon_k,t)$;\\
$\underline{q}_k(t)=\underline{q}_k(\underline{q}_{k-1},u_k+\underline{\epsilon}_k,t)$; $\overline{q}_k(t)=\overline{q}_k(\overline{q}_{k-1},u_k+\overline{\epsilon}_k,t)$;\\
$\xi_k(t) = \operatorname*{max}_{[x',y',\theta']^T \in \{\underline{q}_k,\overline{q}_k\}}||(x_k,y_k),(x',y')||$;\\
\If{$\exists t \in [(k-1)\Delta t,k\Delta t] \text{ s.t. } D((x_k(t),y_k(t),\xi_k(t))\subseteq [\pi_p]$}{$\Theta_k =\Theta_k \cup \{\pi_p\};$
}
\If{$D((x_k,y_k),\xi_k)\subseteq [\pi_d]$}{$\Theta_k =\Theta_k \cup \{\pi_d\},;$}
\If{$\exists t \in [(k-1)\Delta t,k\Delta t] \text{ s.t. } D((x_k(t),y_k(t),\xi_k(t)) \cap [\pi_u] \neq \emptyset$}{$\Theta_k =\Theta_k \cup \{\pi_u\};$}
$s'=(q_{k}(t),\underline{q}_{k},\overline{q}_{k},\underline{\epsilon}_{k},\overline{\epsilon}_{k},\Theta_{k})$; $P(s,u_k,s')=\frac{1}{n};$ $S=S \cup \{s'\}$;
}}
\end{algorithm}

We generate $S$ and $P$ while building $G_K(q_{init})$ starting from $q_{init}$. Algorithm \ref{alg:algorithm1} takes as inputs a state $s \in S$ corresponding to some state trajectory $q_{k-1}(t)$ and an applied control input $u_k+\epsilon_k$, and generates the new state of the MDP and updates $S$ and $P$.
First, given the end state of $q_{k-1}(t)$ and the applied control input $u_k+\epsilon_k \in W_d$, the state trajectory at stage $k$, $q_k(t)$, is obtained (line 2). Then, using $\underline{q}_{k-1}$ and $\overline{q}_{k-1}$, and the fact that $\epsilon_k \in [\underline{\epsilon}_k,\overline{\epsilon}_k] \in \mathcal{E}$, we obtain $\underline{q}_k$ and $\overline{q}_k$ (line 3). The uncertainty trajectory along $q_k(t)$, $\xi_k(t)$ follows from Eqn. (\ref{eq:unceratiny}) (line 4). Using the conditions stated in Sec. \ref{sec:motion}, the algorithm checks if it can be guaranteed that a pick-up region  is entered (lines $5-6$), that the end state is inside of a drop-off region (lines $7-8$), and if it is possible to enter $R_{unsafe}$ (lines $9-10$) along $q_k(t)$ when the uncertainty trajectory is $\xi_k(t)$.

Finally, the newly generated state, $s'$, is added to $S$ and the transition probability function is updated, i.e., $P(s,u_k,s')=\frac{1}{n}$ (line $11$). This follows from the fact that given a control input $u_k \in U$ the applied control input will be $u_k + \epsilon_k \in W_d$ with probability $\frac{1}{n}$, since $p_{\omega}(\omega(u_k)=u_k+\epsilon_k)=\frac{1}{n}$, $\epsilon_k \in E$ (see the MDP fragment in Fig. \ref{fig:MDPGraphics}). When constructing a state $s$ corresponding to a state trajectory $q_K(t) \in G_K(q_{init})$, i.e., when the termination time is reached, we set $A(s)=\varphi$ with $P(s,\varphi,s)=1$.
\begin{prop}
The model $M$ defined above is a valid MDP, i.e., it satisfies the Markov property and $P$ is a transition probability function.
\end{prop}
{\bf{Proof:}}
The proof follows from the construction of the MDP. Given a current state $s \in S$ and an action $u \in A(s)$, the conditional probability distribution of future states depends only on the current state $s$, not on the sequences of events that preceded it (see Alg. \ref{alg:algorithm1}). Thus, the Markov property holds. In addition, since $\sum_{\epsilon \in E}p_{\omega}(\omega(u)=u+\epsilon)=1$, it follows that $P$ is a valid transition probability function. {$\blacksquare$}

\begin{prop}
Let $s_0s_1\ldots \overline{s_K}$ \footnote[4]{The element under the over-line is repeated infinitely since $A(s_K)=\varphi$ and $P(s_K,\varphi,s_K)=1$.} be the path through the MDP corresponding to a state trajectory $q(t) \in G_K(q_{init})$, $t \in [0,K \Delta t]$, i.e., $s_k=(q_{k}(t),\underline{q}_{k},\overline{q}_{k},\underline{\epsilon}_{k},\overline{\epsilon}_{k},\Theta_{k}) \in S$ is such that $q_k(t)=q(t')$, $t' \in [(k-1)\Delta t, k\Delta t]$, $k=1,\ldots,K$. Also, let $u_k \in U$ be such that $q_k(t)=q_k(q_{k-1},u_k+\epsilon_k,t)$, $k=1,\ldots,K$, where $\epsilon_k \in E$, $\epsilon_k \in [\underline{\epsilon}_{k},\overline{\epsilon}_{k}] \in \mathcal{E}$, and $q_0=q_{init}$.
Then, if word $o_0o_1\ldots \overline{o_K}$, where $o_i=h(s_i)$, $i=0,1,\ldots,K$, satisfies PCTL formula $\phi$, the following holds: (i) $q(t)$ when the uncertainty trajectory is $\xi(t)$, such that $\xi(t')=\xi_k(t)$, $t' \in [(k-1)\Delta t, k\Delta t]$, $k=1,\ldots,K$, where $\xi_k(t)$ is given by Eqn. (\ref{eq:unceratiny}), satisfies $\phi$, and (ii) any state trajectory of the original system $q'(t')=q_k'(t)$, $t' \in [(k-1)\Delta t, k\Delta t]$, $k=1,\ldots,K$, such that $q_k'(t)=q_k'(q_{k-1}',u_k+\epsilon_k',t)$, where $\epsilon_k' \in [\underline{\epsilon}_{k},\overline{\epsilon}_{k}] \in \mathcal{E}$ and $q_0'=q_{init}$, satisfies $\phi$. 
\end{prop}
{\bf{Proof:}} For part (i) the proof follows from the conditions stated in Sec. \ref{sec:motion} and from the construction of the MDP. For part (ii) note that due to the conservative approximation of the uncertainty region, $q'(t) \subseteq D((x(t),y(t)),\xi(t))$, $\forall t \in [0,K \Delta t]$. Thus, it follows from (i) that $q'(t)$ satisfies $\phi$.  {$\blacksquare$}

\section{Vehicle Control Strategy}
\label{sec:control}
\subsection{PCTL control policy generation}

We use the PCTL control synthesis approach from \cite{LWAB10} to generate a 
control policy for the MDP $M$. The tool takes as input an MDP and a PCTL formula $\phi$ and returns the control policy that maximizes the probability of satisfying $\phi$, denoted $\mu$, as well as the corresponding probability value, denoted $V$, where $V: S \rightarrow [0,1]$. Specifically, for $s \in S$, $\mu(s) \in A(s)$ is the action to be applied at $s$ and $V(s)$ is the probability of satisfying the specification at $s$ under control policy $\mu$. The tool is based on the off-the-shelf PCTL model-checking tool PRISM (see \cite{kwiatkowska:probabilistic}). 
We use Matlab to construct MDP $M$, which together with $\phi$ is passed to the PCTL control synthesis tool. The computational complexity of this step is as follows: Given $U$, $E$ and $K$, the size of the MDP $M$ is bounded above by $(|U| \times |E|)^{K}$. The time complexity of the control synthesis algorithm is polynomial in the size of the MDP and linear in the number of temporal operators in the formula.

\subsection{Obtaining a vehicle control strategy}
Let  $[\underline{w}_1,\overline{w}_1][\underline{w}_2,\overline{w}_2]\ldots[\underline{w}_{k},\overline{w}_{k}]$ be a sequence of measured intervals, where $[\underline{w}_i,\overline{w}_i]=[u_i+\underline{\epsilon}_i,u_i+\overline{\epsilon}_i]$, $u_i \in U$ and $[\underline{\epsilon}_i,\overline{\epsilon}_i] \in \mathcal{E}$, $i=1,\ldots,k$. 
This corresponds to a unique path through the MDP: $s_0 \xrightarrow{u_1,[\underline{\epsilon}_1,\overline{\epsilon}_1]}s_1 \xrightarrow{u_2,[\underline{\epsilon}_2,\overline{\epsilon}_2]}s_2\ldots{s_{k-1}} \xrightarrow{u_k,[\underline{\epsilon}_k,\overline{\epsilon}_k]}s_{k}$, where each transition is induced by a choice of action and the noise interval. The uniqueness follows from the construction of MDP $M$ and the fact that the noise interval is an element of a state. Given a sequence of measured intervals, we define a mapping function in the form of a finite sequence $\Psi=\{\psi_1,\ldots,\psi_K\}$ where $\psi_k:({U}\times \mathcal{E})^k \rightarrow S$, s.t.
\begin{equation*}
\psi_k((u_1,[\underline{\epsilon}_1,\overline{\epsilon}_1]) \ldots (u_k,[\underline{\epsilon}_k,\overline{\epsilon}_k]))=s_{k}
\end{equation*}
for $k=1,\ldots,K$.

The desired  vehicle control strategy is in the form of a finite sequence $\Gamma= \{\gamma_0,\gamma_1,\ldots \gamma_{K-1}\}$, where $\gamma_0=\mu(s_0) \in U$ and $\gamma_k:({U}\times \mathcal{E})^k \rightarrow U$, s.t.
\begin{eqnarray*}
\lefteqn{\gamma_k((u_1,[\underline{\epsilon}_1,\overline{\epsilon}_1]) \ldots (u_k,[\underline{\epsilon}_k,\overline{\epsilon}_k]))=}\\
&& \mu(\psi_k((u_1,[\underline{\epsilon}_1,\overline{\epsilon}_1]) \ldots (u_k,[\underline{\epsilon}_k,\overline{\epsilon}_{k}])))=\mu(s_{k})
\end{eqnarray*}
for $k=1,\ldots,K-1$. 
At stage $k$, the control input is $u_k=\gamma_{k-1}((u_1,[\underline{\epsilon}_1,\overline{\epsilon}_1]) \ldots (u_{k-1},[\underline{\epsilon}_{k-1},\overline{\epsilon}_{k-1}])) \in U$. Thus, given a sequence of measured intervals, $\Gamma$ returns the control input for the next stage by mapping the sequence to the state of the MDP; the control input corresponds to the optimal action at that state.

\vspace{1mm}
\begin{thm}
\label{thm:theorem1}
The probability that the system given by Eqn. (\ref{eq:system2}), under the control strategy $\Gamma$, generates a trajectory that satisfies PCTL formula $\phi$ (Eqn. (\ref{eq:phi})) is bounded from below by $V(s_0)$, where $V(s_0)$ is the probability of satisfying $\phi$ on the MDP, under the control policy $\mu$.
\end{thm}
\vspace{1mm}

{\bf{Proof:}}
Considering the original system, let $u_k+\epsilon_k$, $k=1,\ldots,K$, be an applied control input at stage $k$, such that $u_k=\gamma_{k-1}((u_1,[\underline{\epsilon}_1,\overline{\epsilon}_1]) \ldots (u_{k-1},[\underline{\epsilon}_{k-1},\overline{\epsilon}_{k-1}])) \in U$ and $\epsilon_k \in [-\epsilon_{max},\epsilon_{max}]$. Then, $[u_k+\underline{\epsilon}_{k},u_k+\overline{\epsilon}_{k}]$  is the measured interval at stage $k$, such that $\epsilon_k \in [\underline{\epsilon}_k,\overline{\epsilon}_k] \in \mathcal{E}$. Let $q(t)$, $t \in[0, K \Delta t]$, be the resulting state trajectory of the original system with $q(0)=q_{init}$.

The sequence of  measured intervals corresponds to a unique path through the MDP, denoted $s_0s_1\ldots \overline{s_{K}}$, where $s_k=\psi_{k}((u_1,[\underline{\epsilon}_1,\overline{\epsilon}_1])\ldots (u_{k},[\underline{\epsilon}_{k}\overline{\epsilon}_{k}]))$, $k=1,\ldots,K$. This sequence of states produces word $o=o_0o_1 \ldots \overline{o_{K}}$, where $o_i=h(s_i)$, $i=0,1,\ldots,K$.  The produced word can: (i) satisfy $\phi$ and (ii) not satisfy $\phi$. Let us first consider the former.

If the word satisfies $\phi$ from Proposition 2 it follows that $q(t)$ also satisfies $\phi$. Under $\Gamma$, the probability of generating a state trajectory such that $\epsilon_k \in [\underline{\epsilon}_k,\overline{\epsilon}_k]$, $k=1,\ldots,K$, is equivalent to the probability of generating $s_0s_1\ldots \overline{s_{K}}$ under $\mu$. Since under $\mu$ the probability that the MDP generates a satisfying word is $V(s_0)$ it follows that the probability that the original system under $\Gamma$ generates a satisfying trajectory is also $V(s_0)$.

To show that $V(s_0)$ is the lower bound we need to consider the latter case. It is sufficient to observe that because of the conservative approximation of the uncertainty region it is possible that $q(t)$ satisfies $\phi$, even though word $o$ does not satisfy it.
Therefore, it follows that the probability that the original system, under the optimal control startegy $\Gamma$, will generate a satisfying trajectory is bounded from below by $V(s_0).$ As a final remark, note that the bound obtained in this work approaches the true probability of satisfying $\phi$ in the theoretical limit, as $\Delta \epsilon \rightarrow 0$.
$\blacksquare$

\section{Case study}
\label{sec:casestudy}
We considered the system given by Eqn. (\ref{eq:system2}) and we used the following numerical values: $1/\rho=\pi/3$, $\Delta t=1.2$, and $\epsilon_{max}=0.06$ with $n=3$, i.e., $\Delta \epsilon=0.04$. Thus, the maximum actuator noise was approximately $6\%$ of the maximum control input.

\begin{figure}
\begin{center}
\includegraphics[width=0.478\textwidth]{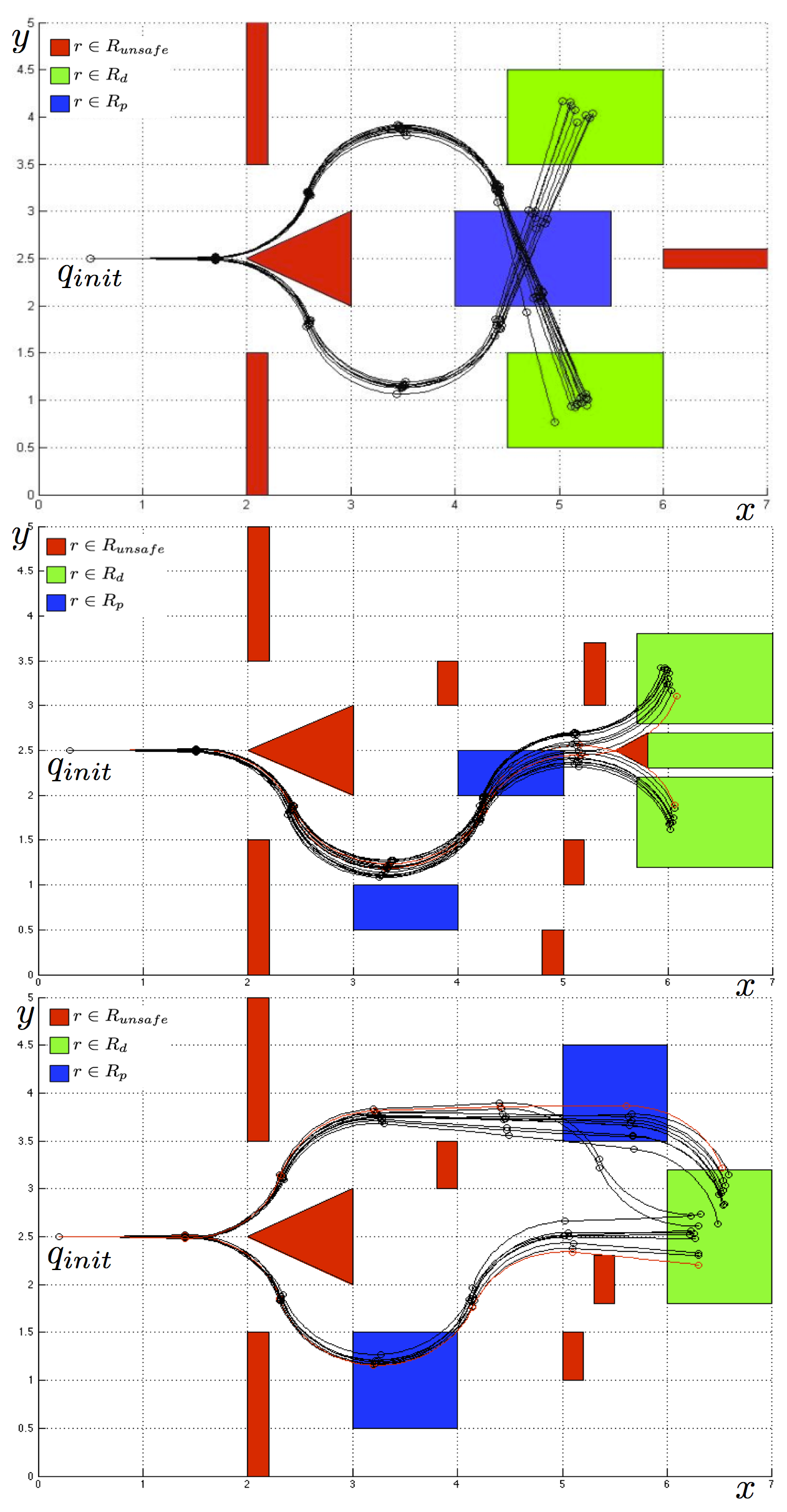}
\caption{20 sample state (position) trajectories for cases $A$, $B$, and $C$ (to be read top-to-bottom). The unsafe, pick-up, and the drop-off regions are shown in red, blue and green, respectively. Satisfying and violating trajectories are shown in black and red, respectively. In case $A$, all state trajectories were satisfying. }
\label{fig:Runs}
\end{center}
\end{figure}

Three case studies are shown in Fig. \ref{fig:Runs}. 
The maximum probability of satisfying PCTL formula $\phi$ (Eqn. (\ref {eq:phi})) on the MDPs corresponding to cases $A$, $B$ and $C$ are $0.981$, $0.874$ and $0.892$, respectively. 
For all three case studies, we found that $K=6$ was enough as a terminal time, 
and we found the vehicle control strategies through the method described in Sec. \ref{sec:control}. To verify our result, we simulated the original system under the obtained vehicle control strategies. 

In Fig. \ref{fig:Runs}, we show sample state trajectories and in Table 1 we compare the satisfaction probabilities obtained on the MDP with the simulation based satisfaction probabilities (number of satisfying trajectories over the number of generated trajectories). The results support Theorem \ref{thm:theorem1}, since the simulation based probabilities are bounded from below by the theoretical probabilities. 

\begin{table}[h]
\caption{Theoretical and simulation based probabilities of satisfying the specification.}
\label{table:data}
\begin{center}
\scalebox{1}{
\begin{tabular}{| c || c || c || c || c |}
\hline 
Case & Theoretical Probability & \multicolumn{3}{c|}{Number of generated trajectories}\\ 
\cline {3-5}&  $(V(s_0))$ &  $10^3$ &  $5 \cdot 10^3$ & $10^4$\\
\cline {3-5}& & \multicolumn{3}{c|}{Probabilities from simulations}\\
\hline
$A$ & $0.981$ & $1$ & $1$ & $1$\\
\hline
$B$ & $0.874$ & $0.902$ & $0.926$ & $0.928$ \\
\hline
$C$ & $0.892$ & $0.913$ & $0.931$ & $0.935$\\
\hline
\end{tabular}
}
\end{center}
\end{table}

For each case study, the constructed MDP had approximately $20 000$ states. The Matlab code used to construct the MDP ran for approximately $4$ minutes on a computer with a 2.5GHz dual processor. The control synthesis tool generated an optimal policy in about $1$ minute. 

\section{Conclusion and future work}
\label{sec:conclusion}

We developed a feedback control strategy for a stochastic Dubins vehicle
such that the probability of satisfying a temporal logic statement over some environmental properties is maximized. Through discretization and quantization, we translated this problem to finding a control policy maximizing the probability of satisfying a PCTL formula on an MDP. 
We showed that the probability that the vehicle satisfies the specification in the original environment is bounded from below by the maximum probability of satisfying the specification on the MDP.

Future work includes extensions of this approach to controlling different types of vehicle models (e.g., stochastic car-like vehicles with uncertainties in both the forward speed and the turning rate), allowing for richer temporal logic specifications, and experimental validations.   

\bibliographystyle{alpha} 
\bibliography{iros2012}   

\end{document}